%% file: main.tex
\documentclass{article}
\usepackage{PRIMEarxiv}
\usepackage[T1]{fontenc}
\usepackage{microtype}
\usepackage[numbers,square,sort&compress]{natbib}
\input{math_commands.tex}

\usepackage{booktabs}
\usepackage{multirow}
\usepackage[table]{xcolor}
\definecolor{textualrlblue}{RGB}{234,244,252}
\usepackage{graphicx}
\usepackage{float}
\usepackage{capt-of}
\usepackage{needspace}
\usepackage{algorithm}
\usepackage{algpseudocode}
\usepackage{hyperref}
\usepackage{url}
\usepackage{enumitem}
\usepackage[breakable]{tcolorbox}


\title{TextualRL: Textual Reinforcement Learning\\for Context Management}
\author{
  Tianyu Huai$^{1,2}$, Jie Zhou$^1$\thanks{Corresponding author.}, Bo Mao$^1$, Qin Chen$^1$, Liang He$^1$, Xipeng Qiu$^{2,3}$ \\
  $^1$ School of Computer Science and Technology, East China Normal University, Shanghai\\
  $^2$ Shanghai Innovation Institute, $^3$ Fudan University \\
  \texttt{tyhuai@stu.ecnu.edu.cn, jzhou@cs.ecnu.edu.cn} \\[0.4em]
  \textcolor{red}{\url{https://github.com/ECNU-ICALK/TextualRL}}
}
\hypersetup{hidelinks,pdftitle={TextualRL: Textual Reinforcement Learning for Context Management},pdfauthor={Tianyu Huai, Jie Zhou, Bo Mao, Qin Chen, Liang He, Xipeng Qiu}}

\begin{document}
\maketitle
\begin{abstract}
\input{sections/abstract}
\end{abstract}

\input{sections/introduction}
\input{sections/related_work}
\input{sections/method}
\input{sections/experiments}
\input{sections/conclusion}

\bibliographystyle{unsrtnat}
\bibliography{references}

\clearpage
\appendix
\input{sections/appendix}
\end{document}

%% file: math_commands.tex
\usepackage{amsmath,amsfonts,bm}

\def\eqref#1{equation~\ref{#1}}

\def\1{\bm{1}}

\DeclareMathAlphabet{\mathsfit}{\encodingdefault}{\sfdefault}{m}{sl}
\SetMathAlphabet{\mathsfit}{bold}{\encodingdefault}{\sfdefault}{bx}{n}



%% file: sections/abstract.tex
LLM agents can adapt to task feedback by revising the textual context that guides their behavior, without changing their model parameters. However, turning execution rewards into reusable context updates is challenging: individual trajectories offer limited comparative evidence about what to change, while task-specific fixes may overgeneralize when incorporated into a shared context. We introduce TextualRL, a gradient-free reinforcement learning framework that treats a group of rollouts from the same task under the same context, rather than an isolated trajectory, as the unit of policy critique. \emph{Groupwise Policy Critique} contrasts successful and failed rollouts within a task; when outcomes are uniform, it draws on cross-task evidence to identify behaviors worth preserving or formulate repair hypotheses. \emph{Cross-Group Policy Update} reviews the resulting proposals against evidence pooled across groups, including observations that yield no proposed edit, to refine their content and applicability. Reviewed proposals form a bounded textual patch, with adoption of the candidate context determined by held-out validation rewards. Together, these components establish a reward-driven policy optimization loop in which multiple attempts jointly inform reusable context updates. We evaluate TextualRL with five target agents on six benchmarks. Averaged over the six benchmarks, TextualRL improves test accuracy over the base model by 24.4 and 15.3 percentage points for GPT-5.5 and Qwen3.8-27B, respectively.

%% file: sections/introduction.tex
\section{Introduction}
\label{sec:introduction}

Reinforcement learning (RL) improves an agent's behavior through interaction and reward~\citep{kaelbling1996rlsurvey}. For LLM agents whose parameters are unavailable for optimization, textual context offers an alternative interface for adaptation: instructions and procedures condition how a fixed model interprets tasks, uses tools, and produces outputs. Reflexion~\citep{shinn2023reflexion} incorporates verbal feedback into subsequent attempts, while Agentic Context Engineering (ACE)~\citep{zhang2026ace} evolves contextual playbooks from execution experience. These developments motivate policy optimization in context space: the objective remains higher task reward, but the optimization variable is the context rather than the model weights.

Existing work provides several mechanisms for optimizing textual artifacts. ProTeGi~\citep{pryzant2023protegi} and TextGrad~\citep{yuksekgonul2025textgrad} use natural-language feedback to guide updates, while GEPA~\citep{agrawal2026gepa} combines reflection on execution traces with evolutionary prompt search. For agent skills, Trace2Skill~\citep{ni2026trace2skill} extracts trajectory-local lessons and hierarchically consolidates them into reusable instructions. SkillOpt~\citep{yang2026skillopt} instead analyzes successful and failed trajectory minibatches separately, merges their proposed edits, and validates bounded skill updates. These methods establish the value of execution-driven adaptation, while highlighting an important design choice: how should trajectories be organized before their feedback is converted into a shared context update?

Group-based RL provides a useful organizing principle. Group Relative Policy Optimization (GRPO)~\citep{shao2024deepseekmath} samples multiple responses to the same question and uses their relative rewards to guide parameter updates. For textual adaptation, a group of same-task rollouts likewise provides a comparison unavailable from an isolated execution: under the same task and context, which behavioral differences accompany different outcomes? Comparative experience learning is also explored by ExpeL~\citep{zhao2024expel} and ReasoningBank's parallel MaTTS~\citep{ouyang2026reasoningbank}, which extract reusable insights or memories from alternative executions. Building on this perspective, we ask:
\emph{how can groups of rewarded executions drive policy optimization when updates are made to shared textual context rather than model parameters?} The goal is not merely to select a successful attempt, but to use multiple attempts to improve the policy governing future executions.

Turning grouped experience into a context update presents two challenges. First, rewards identify outcomes, not the instructions that should change. Within-task success--failure comparisons provide evidence about alternative behaviors, but this contrast disappears when every sampled rollout succeeds or every rollout fails. An analysis procedure must therefore distinguish evidence supporting preservation from evidence suggesting an unverified repair. Second, a context is shared across tasks, whereas a proposed modification may be supported only under particular conditions. Merging proposals alone cannot reveal every exception: a successful execution may require no edit while still contradicting another group's
proposed rule. Effective updates therefore require both outcome-aware policy critique and a review of the evidence that supports or limits each modification. 

\begin{figure}[!t]
    \centering
    \includegraphics[width=0.95\linewidth,pagebox=cropbox]{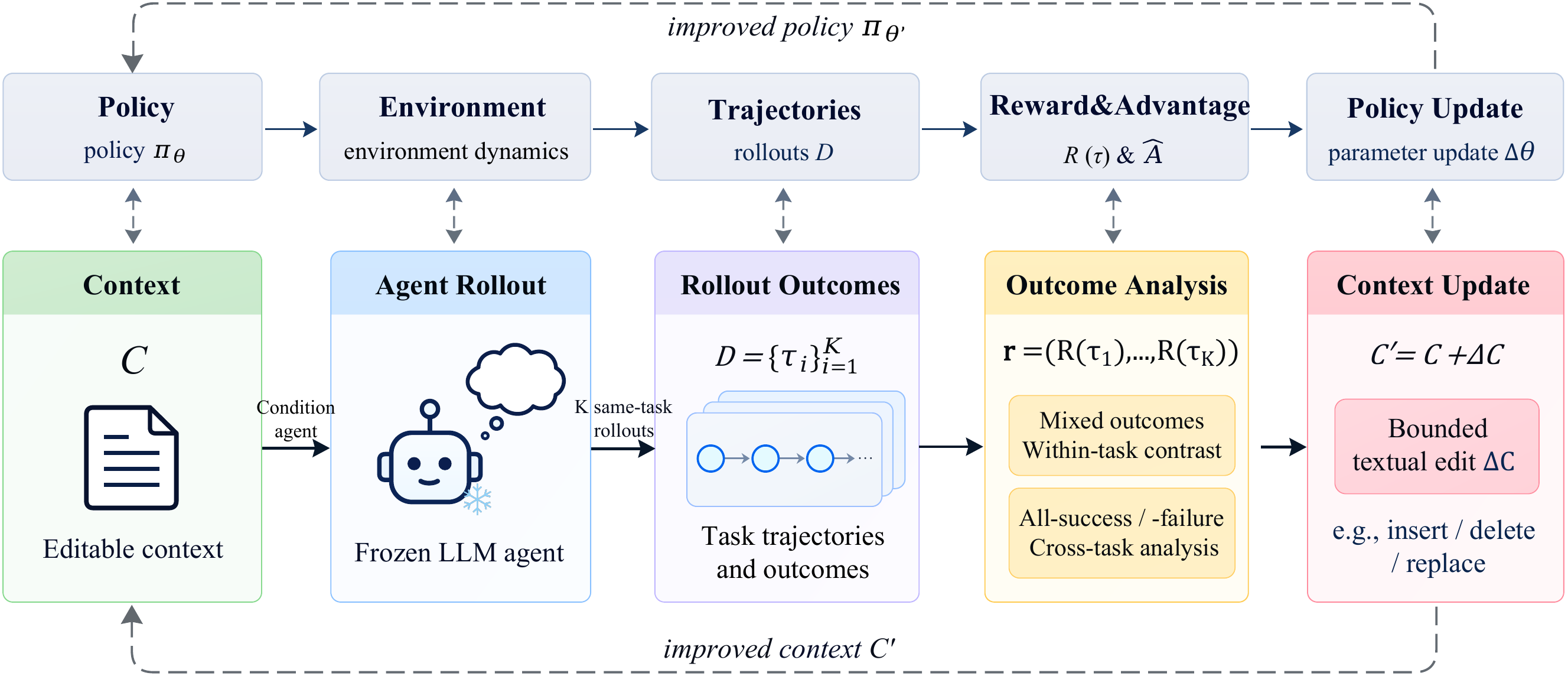}
    \vspace{-4mm}
    \caption{
        Conceptual comparison of parameter-space and context-space policy optimization. TextualRL uses groups of same-task rollouts and their rewards to propose textual edits $\Delta C$ to context $C$, while the target model remains frozen. 
    }
    \label{fig:intro}
    \vspace{-4mm}
\end{figure}

We introduce TextualRL, a gradient-free reinforcement learning algorithm for context management that treats same-task rollout groups as the basic unit of policy critique. As illustrated in Figure~\ref{fig:intro}, each iteration collects multiple executions per task under the current context while keeping the target model frozen. \emph{Groupwise Policy Critique} compares successful and failed rollouts within mixed-outcome tasks; for all-success or all-failure tasks, it uses cross-task evidence to identify behaviors worth preserving or formulate repair hypotheses, respectively.
The critics return edit proposals and trajectory evidence separately, retaining observations even when no edit is proposed. \emph{Cross-Group Policy Update} reviews consolidated proposals against evidence pooled across analysis groups, retaining, narrowing, or revising instructions according to their support and applicability. The reviewed edits form a bounded textual patch. A step-level validation gate accepts the candidate context only if its reward on held-out tasks exceeds the recorded reference score. The selected context then conditions subsequent rollouts, closing a reward-driven policy optimization loop without model-parameter updates.
We evaluate TextualRL with five target agents across six benchmarks. Averaged over the six benchmarks, TextualRL improves test accuracy over the no-skill baseline by 24.4 and 15.3 percentage points for GPT-5.5 and Qwen3.8-27B, respectively. The corresponding gains over the best baseline are 0.9 and 3.2 percentage points.

Our contributions are threefold:
\begin{itemize}[leftmargin=*, align=left]
    \item We introduce TextualRL, a group-based algorithm for gradient-free policy optimization over the shared textual context of a frozen LLM agent.
    \item We develop Groupwise Policy Critique, which uses the available reward structure to combine within-task comparisons with cross-task preservation and repair analysis.
    \item We develop Cross-Group Policy Update, which retains evidence independently of edit proposals to review their content and applicability before bounded, validation-gated updates.
\end{itemize}

%% file: sections/related_work.tex
\section{Related Work}
\label{sec:related_work}

\paragraph{Prompt and Context Optimization.}
Automatic prompt optimization searches for text that improves task performance. AutoPrompt~\citep{shin2020autoprompt} uses gradient-guided search over discrete trigger tokens, whereas APE~\citep{zhou2023ape} generates and evaluates instructions from demonstrations. DSPy~\citep{khattab2024dspy} compiles modular language-model pipelines through demonstration bootstrapping and selection, while MIPRO~\citep{opsahlong2024mipro} jointly optimizes instructions and demonstrations across modules. Context and skill optimization extends this scope to reusable guidance for future executions. Dynamic Cheatsheet~\citep{suzgun2026dynamiccheatsheet} maintains adaptive memory across test queries. Trace2Skill~\citep{ni2026trace2skill} independently extracts trajectory-local patches and hierarchically consolidates them into agent skills. EvoSkill~\citep{alzubi2026evoskill} proposes skills from failed trajectories and evaluates candidate skill repositories through validation. SkillOpt~\citep{yang2026skillopt} separately analyzes successful and failed trajectory minibatches, merges their proposals, and selects bounded skill updates using held-out validation. TextualRL shares the goal of feedback-driven textual adaptation but changes how executions inform an update: same-task rollouts under the same context are preserved for joint critique before proposal consolidation. Moreover, trajectory evidence is retained independently of proposed edits. This allows observations that warrant no new instruction to constrain other proposals, making evidence about a rule's applicability available beyond the editing suggestions themselves.

\paragraph{Learning from Trajectory Feedback.}
Trajectory feedback supports both task execution and learning across attempts. ReAct~\citep{yao2023react} interleaves reasoning and actions with environment observations, and Self-Refine~\citep{madaan2023selfrefine} iteratively critiques and revises an output. LATS~\citep{zhou2024lats} combines tree search, self-reflection, and environment feedback to explore alternative trajectories. Voyager~\citep{wang2024voyager} uses execution feedback to improve programs and accumulate reusable code skills. Comparative experience learning is also established: ExpeL~\citep{zhao2024expel} extracts insights from same-task success--failure pairs and shared patterns across successful tasks, while ReasoningBank~\citep{ouyang2026reasoningbank} uses parallel MaTTS to curate reusable memory by comparing multiple executions of the same query. In parameter-space RL, GRPO~\citep{shao2024deepseekmath} uses relative rewards within groups of same-question responses to guide policy updates. TextualRL instead performs gradient-free policy optimization over a shared context, coupling comparative experience with an explicit proposal--review--validation loop. \emph{Groupwise Policy Critique} adapts its analysis to the available reward structure: mixed outcomes support within-task contrast, whereas uniform outcomes require cross-task evidence for preservation or repair hypotheses. \emph{Cross-Group Policy Update} then reviews both the content and applicability of proposed instructions using independently pooled evidence, before edits undergo held-out validation. Together, these mechanisms connect outcome-aware comparison to cross-task evidence review, turning groups of attempts into reward-selected updates to a frozen agent's shared policy context.

%% file: sections/method.tex
\section{TextualRL}
\label{sec:method}

\subsection{Context-Space Policy Optimization}
\label{sec:context-objective}
\label{sec:overview}

Let $C\in\mathcal{C}$ denote the editable textual context containing instructions and procedures shared across tasks. For a task $x$, the context induces a trajectory distribution $\pi_\theta(\tau\mid x,C)$ through the target agent, where a trajectory $\tau$ comprises actions, available observations, and the final output. The target parameters $\theta$ remain fixed, and TextualRL optimizes the context to maximize expected task reward:
\begin{equation}
    \max_{C\in\mathcal{C}} J(C)
    = \mathbb{E}_{x\sim p_{\mathrm{task}}}
      \mathbb{E}_{\tau\sim\pi_\theta(\cdot\mid x,C)}[R(\tau)],
    \label{eq:objective}
\end{equation}
where $p_{\mathrm{task}}$ is the task distribution and $R(\tau)\in\{0,1\}$ indicates task success. Policy optimization thus operates through textual changes to $C$ rather than updates to $\theta$.

At step $t$, we sample a training batch $B_t$ and collect $K$ rollouts per task under the current context $C_t$. For each $x\in B_t$, we retain its indexed rollout group and reward vector:
\begin{equation}
    D_x=\{\tau_{x,i}\}_{i=1}^{K},
    \qquad
    \mathbf{r}_x=(r_{x,1},\ldots,r_{x,K}),
    \label{eq:task-block}
\end{equation}
where $r_{x,i}=R(\tau_{x,i})$. We omit the step index from $D_x$ and $\mathbf{r}_x$ for clarity. Keeping same-task rollouts together preserves the behavioral and outcome comparisons used to propose an update.

\begin{figure}[!t]
    \centering
    \includegraphics[width=\linewidth,pagebox=cropbox]{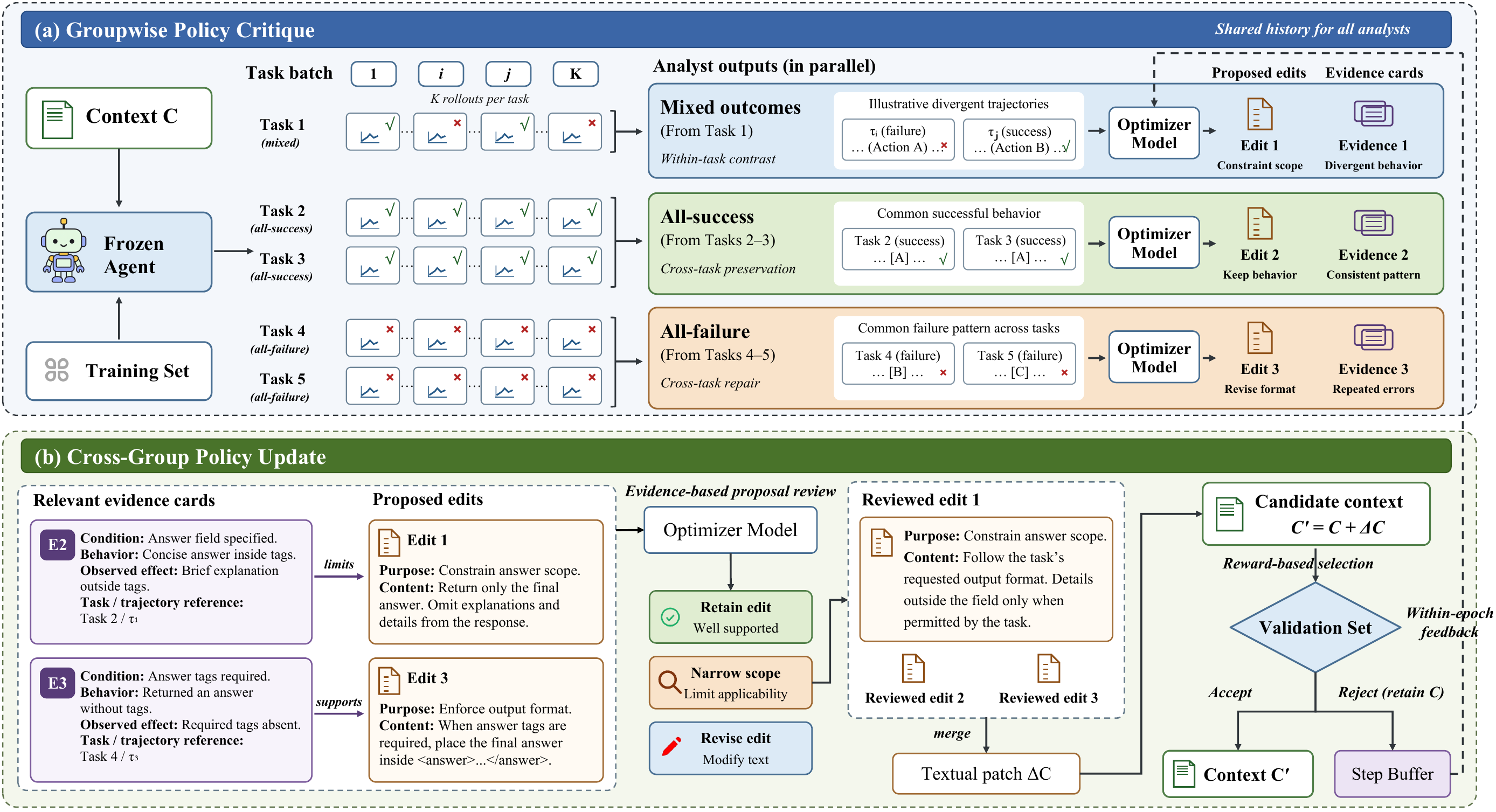}
    \vspace{-4mm}
    \caption{
        Overview of TextualRL. (a) Groupwise Policy Critique preserves same-task rollout groups and uses within-task contrast for mixed outcomes, cross-task preservation for all-success outcomes, and cross-task repair for all-failure outcomes. Parallel critics return edit proposals and evidence cards separately. (b) Cross-Group Policy Update reviews proposals against pooled evidence, constructs a bounded patch, and selects the candidate context through held-out validation. Within-epoch feedback informs subsequent critique calls.
    }
    \label{fig:overview}
    \vspace{-3mm}
\end{figure}

As shown in Figure~\ref{fig:overview}, each step comprises two modules. \emph{Groupwise Policy Critique} uses the available outcome structure to produce edit proposals and trajectory evidence. \emph{Cross-Group Policy Update} reviews these proposals against pooled evidence, constructs a bounded textual patch, and uses held-out validation rewards to select the next context. Subsequent rollouts are conditioned on the selected context, forming a gradient-free policy optimization loop.

\subsection{Groupwise Policy Critique}
\label{sec:outcome-proposals}

The comparison available within a rollout group depends on its reward structure. Let $n_x=\sum_{i=1}^{K}r_{x,i}$ be the number of successful rollouts for task $x$. We route each group according to
\begin{equation}
    \operatorname{route}(\mathbf{r}_x)=
    \begin{cases}
        \text{mixed}, & 0<n_x<K,\\
        \text{all-success}, & n_x=K,\\
        \text{all-failure}, & n_x=0.
    \end{cases}
    \label{eq:outcome-routing}
\end{equation}
Before critique, we consolidate duplicate trajectories while preserving their outcome labels, original rollout references, and multiplicities. All analysis groups retain complete task blocks; multiple rollouts of one task do not count as support from multiple distinct tasks.

\paragraph{Within-task contrast.}
Each mixed-outcome task forms a separate analysis group. The critic compares successful and failed rollouts to identify the earliest observable behavioral divergence, such as a different action or use of available evidence. It proposes at most one atomic repair or preservation rule supported by this contrast, and returns no edit when no supported distinction is observable.

\paragraph{Cross-task preservation.}
When all rollouts succeed, there is no within-task success--failure contrast. We therefore group all-success task blocks into analysis minibatches containing at least two distinct tasks. The critic identifies a shared, observable behavior worth preserving and proposes an instruction only when that behavior is not already covered by $C_t$.

\paragraph{Cross-task repair.}
All-failure task blocks are grouped in the same manner. The critic identifies a shared, observable failure mechanism across at least two distinct tasks and proposes a corresponding repair hypothesis. Failure alone neither supplies a correct answer nor verifies the proposed repair. Both homogeneous routes return no edit when the required common behavior or mechanism is unsupported, and a route with fewer than two task blocks is skipped.

\paragraph{Edit proposals and trajectory evidence.}
Let $g\subseteq B_t$ be the tasks assigned to one critique call and $\mathcal{G}_t$ the collection of these analysis groups. A mixed group contains one task, whereas a homogeneous group contains at least two tasks from the same outcome category. Each call returns proposed edits $P_g$ and evidence cards $E_g$:
\begin{equation}
    (P_g,E_g)=\operatorname{Critique}
    \bigl(C_t,H_t,\{(x,D_x,\mathbf{r}_x):x\in g\}\bigr).
    \label{eq:analyst-outputs}
\end{equation}
Here, $\operatorname{Critique}$ invokes the optimizer model with route-specific instructions. Calls run in parallel and share the current context $C_t$ and within-epoch history $H_t$.

An edit proposal specifies a modification to $C_t$. An evidence card records an observable condition, behavior, and effect, together with a short supporting quote and references to the corresponding task and rollout. Unobserved conditions or effects are marked as unknown. Cards may be returned even when $P_g$ is empty: behavior already handled
correctly by the current context can still constrain a proposal from another group. We therefore pool proposals and evidence independently:
\begin{equation}
    P_t=\bigcup_{g\in\mathcal{G}_t}P_g,
    \qquad
    E_t=\bigcup_{g\in\mathcal{G}_t}E_g.
    \label{eq:proposal-evidence-pools}
\end{equation}

\subsection{Cross-Group Policy Update}
\label{sec:cross-group-evidence}
\label{sec:context-update}

A modification supported by one group may restrict behavior that succeeds elsewhere. Cross-Group Policy Update therefore reviews the content and applicability of local proposals before evaluating their combined effect on the shared context.

\paragraph{Evidence-based proposal review.}
The optimizer first consolidates $P_t$ through hierarchical merging to remove redundancy and resolve conflicting instructions. It then reviews the consolidated candidates using $C_t$ and the independently pooled evidence $E_t$. For each edit, it compares the prescribed behavior and scope with relevant observations, including the behavior being regulated, the failure mechanism being addressed, or an exception within the proposed scope. The review records whether these observations \emph{support}, \emph{limit}, or \emph{contradict} the instruction.

Based on these relations, the optimizer retains the edit, narrows its applicability using an observable condition, or revises its content. The resulting edits form $\widetilde{P}_t$, with each decision accompanied by a rationale and references to the evidence used. Insufficient relevant evidence leaves the proposal's original scope and uncertainty unchanged. The review neither requires evidence from every outcome category nor introduces unrelated edits to account for unused cards.
Figure~\ref{fig:overview}(b) illustrates this distinction. Card~E2 records a successful response with the answer inside the required field and an explanation outside it, limiting Edit~1's blanket answer-only restriction. The revised instruction follows the requested format and allows details outside the field only when permitted by the task. Card~E3 instead supports Edit~3 by documenting missing required answer tags. The pool $E_t$ contains only current-step evidence, so this review requires no additional target-agent rollouts.

\paragraph{Textual patch.}
Each reviewed edit specifies an operation---\texttt{append}, \texttt{insert\_after}, \texttt{replace}, or \texttt{delete}---together with target text when required and replacement or insertion content. We select at most $L_t$ edits, ranking candidates with the optimizer when they exceed this budget. We also remove edits with explicit dependencies on information unavailable during target execution. The remaining operations form a textual patch $\Delta C_t$, yielding
\begin{equation}
    C'_t=C_t+\Delta C_t
    \equiv \operatorname{Apply}(C_t,\Delta C_t),
    \label{eq:textual-update}
\end{equation}
where $+$ denotes applying textual operations rather than numerical addition.

\paragraph{Reward-based selection.}
We evaluate $C'_t$ on a held-out validation set $\mathcal{V}$ using the benchmark's binary success metric. Let $\widehat{V}(C)$ denote mean validation success under the evaluation protocol and $s_t$ the trainer's recorded reference score. The step-level selection rule is
\begin{equation}
    (C_{t+1},s_{t+1})=
    \begin{cases}
        (C'_t,\widehat{V}(C'_t)), & \widehat{V}(C'_t)>s_t,\\
        (C_t,s_t), & \text{otherwise}.
    \end{cases}
    \label{eq:validation-selection}
\end{equation}
Thus, an accepted candidate updates both the active context and the reference score; a rejected candidate leaves both unchanged. Validation evaluates the complete candidate context, not the individual effects of its constituent edits.

\paragraph{Within-epoch feedback.}
After validation, the shared step buffer augments $H_t$ with observed rollout failure patterns. For rejected candidates, it also records the attempted edit combination and its validation score. Subsequent critique calls in the same epoch receive this history to reduce repeated ineffective proposals. Rejection concerns the evaluated combination and does not imply that every constituent edit is harmful. This history is distinct from $E_t$, which contains only the current step's evidence. Algorithm~\ref{alg:textualrl-step} summarizes the step-level procedure. Appendix~\ref{app:optimizer-prompts} provides the optimizer prompt blocks.

%% file: sections/experiments.tex
\input{sections/main_results_table}

\section{Experiments}
\label{sec:experiments}

\subsection{Experimental Setups}
\paragraph{Benchmarks.}

We evaluate TextualRL on six benchmarks spanning question answering, document understanding, mathematical reasoning, and interactive task execution. SearchQA~\citep{dunn2017searchqa} requires answering questions from provided search snippets. Spreadsheet~\citep{ma2024spreadsheetbench} evaluates workbook manipulation through executable Python programs. OfficeQA~\citep{opsahlong2026officeqa} requires reasoning over enterprise documents using local file tools and supplied source-page evidence. DocVQA~\citep{mathew2021docvqa} pairs questions with document images. LiveMathematicianBench~\citep{he2026livemathematicianbench}, abbreviated as LiveMath, tests mathematical reasoning through multiple-choice questions. ALFWorld~\citep{shridhar2021alfworld} evaluates sequences of textual actions for completing household tasks. We use the released benchmark splits from SkillOpt~\citep{yang2026skillopt}, with dataset details, split sizes, and evaluation metrics provided in Appendix~\ref{app:benchmarks}. The training split supplies trajectories for context optimization, while validation scores determine step-level update acceptance and checkpoint selection. The test split is used only to evaluate the selected contexts.

\paragraph{Baselines.}
We compare TextualRL with seven baselines in Table~\ref{tab:main-results}. \textbf{Base Model} evaluates the frozen target using the original LLM. \textbf{Human skill} uses manually written task guidance, while \textbf{LLM skill} uses a skill generated in a single pass. These baselines measure the benefit of adding static instructions without iterative optimization. \textbf{Trace2Skill}~\citep{ni2026trace2skill} extracts lessons from individual training trajectories and consolidates them into skills. \textbf{TextGrad}~\citep{yuksekgonul2025textgrad} updates text using language-model feedback as textual gradients. \textbf{GEPA}~\citep{agrawal2026gepa} uses reflective evolutionary search to improve prompts from execution feedback. \textbf{SkillOpt}~\citep{yang2026skillopt} aggregates edit proposals from minibatches and selects bounded skill updates through validation. These main comparisons use the benchmark-specific direct-chat setting, including the tools or environment interactions required by each task.

\paragraph{Implementation Details.}
We use five frozen target models: GPT-5.5, GPT-5.4, Qwen3.5-4B, Qwen3.6-35B-A3B, and Qwen3.8-27B. A separate context is optimized for every target--benchmark pair. Unless otherwise specified, GPT-5.5 serves as the optimizer. The target model's parameters remain unchanged during both optimization and evaluation.
Starting from the benchmark-specific initial context, we optimize for four epochs with up to 40 tasks per batch and $K=4$ rollouts per task. A full batch therefore contains 160 trajectories. Analysis and merge minibatches have size eight, subject to the task-group constraints in Section~\ref{sec:outcome-proposals}. The edit budget follows a cosine schedule from four to two edits. The within-epoch buffer provides feedback to subsequent critique calls. Additional implementation details appear in Appendix~\ref{app:implementation}.

\subsection{Main Results}
\label{sec:main-results}

Table~\ref{tab:main-results} compares TextualRL with the seven baselines across all five target models. TextualRL achieves the highest six-benchmark average for every target, with gains of 11.4--24.4 percentage points over Base Model and 0.9--3.2 points over the strongest baseline SkillOpt. It outperforms all baselines in 29 of the 30 target--benchmark settings and ties SkillOpt on GPT-5.5 with ALFWorld. For GPT-5.5, the average score increases from 58.8\% without a skill to 83.2\% with TextualRL. For Qwen3.8-27B, the corresponding scores are 55.6\% and 71.0\%. The improvements therefore extend across both GPT and Qwen targets, with a separately optimized context for each target and benchmark. 
Specifically, TextualRL raises Qwen3.5-4B from 30.6\% without a skill to 82.1\% on ALFWorld, a gain of 51.5 percentage points and the largest improvement over Base Model. This result shows that optimizing context can substantially improve task completion even for the smallest frozen target. On Qwen3.8-27B, TextualRL reaches 56.0\% on LiveMath and 60.7\% on Spreadsheet, exceeding the strongest baseline by 8.0 and 4.3 points, respectively. The advantage is more modest on DocVQA, ranging from 0.3 to 1.4 points over the strongest baseline across the five targets.

\subsection{Ablations}
\label{sec:ablations}

\input{sections/ablation_results_table}

To evaluate the contribution of each module, we conduct ablation studies on Qwen3.8-27B and report the results in Table~\ref{tab:ablations}. For Groupwise Policy Critique, \emph{w/o outcome routing} pools all trajectories collected in a training step and partitions them directly into analyst minibatches of eight trajectories, without first separating them into mixed-outcome, all-success, and all-failure groups. This compares outcome-aware critique with direct minibatch analysis. We also evaluate a single rollout per task ($K=1$). For Cross-Group Policy Update, \emph{intra-group cards only} restricts the review of each merged edit to evidence cards from its originating critique groups. We additionally remove cards produced without edit proposals or disable cross-group proposal review.

The results show that removing outcome routing reduces the six-benchmark average by 1.5 percentage points, while using $K=1$ reduces it by 2.5 points. These results support the value of both organizing critique by outcome patterns and collecting multiple rollouts per task in Groupwise Policy Critique. For Cross-Group Policy Update, restricting review to cards from the originating critique groups reduces the average by 1.2 points, while removing cards without edits reduces it by 1.7 points. These results support using evidence beyond a proposal's originating groups and retaining observations that do not directly produce an edit. Disabling cross-group proposal review reduces the average by 2.7 points, highlighting the value of using evidence to assess and refine proposals before updating the context. Full TextualRL achieves the highest score on every benchmark in both groups.

\subsection{Additional Experiments}
\label{sec:additional-experiments}

\input{sections/run_stability_table}

\paragraph{Run stability.}
\label{sec:run-stability}

To assess variation across runs, we repeat context optimization three times for TextualRL and SkillOpt on Qwen3.8-27B, keeping the data splits and evaluation protocol fixed. Each run independently selects its checkpoint by validation score before test evaluation. Table~\ref{tab:run-stability} reports the mean and sample standard deviation across runs. TextualRL achieves an overall mean of 70.68\%, compared with 67.43\% for SkillOpt, and has a higher mean score on all six benchmarks. The largest mean gain is on LiveMath, at 8.27 points. The overall advantage ranges from 3.18 to 3.36 points across the three runs, showing that the improvement persists across repeated optimization runs.

\input{sections/rollout_count_table}

\paragraph{Rollout count.}
\label{sec:rollout-count}

To examine the effect of trajectory sampling, we vary the number of rollouts per training task over $K\in\{1,2,4,8\}$ on Qwen3.8-27B. Table~\ref{tab:rollout-count} shows that the overall score increases from 68.4\% at $K=1$ to 71.0\% at $K=4$. Increasing $K$ from four to eight doubles the rollout count per task but adds only 0.1 percentage points to the overall score. The additional rollouts do not improve every benchmark: Spreadsheet and DocVQA decrease by 0.7 and 0.3 points, respectively, while LiveMath remains unchanged. In this setting, $K=4$ captures most of the observed improvement with half the rollouts per task required by $K=8$, so we use it as the default setting.

\input{sections/harness_results_table}

\paragraph{Agent harnesses.}
\label{sec:harness-results}

To evaluate TextualRL beyond direct chat, we use GPT-5.5 within the Codex and Claude Code harnesses on five benchmarks, excluding ALFWorld. Table~\ref{tab:harness-results} includes the harness baselines reported by SkillOpt, with EvoSkill~\citep{alzubi2026evoskill} as an additional skill evolution baseline. TextualRL achieves five-benchmark averages of 80.1\% and 78.6\% under Codex and Claude Code, respectively, improving over Base Model by 26.1 and 20.8 percentage points. Relative to SkillOpt, the average gains are 1.3 and 1.7 points. TextualRL obtains the highest score in nine of the ten harness--benchmark settings. The exception is Codex LiveMath, where TextualRL scores 76.8\% versus SkillOpt's 78.4\%. These results support the effectiveness of context optimization across both agent harnesses.

\input{sections/skill_transfer_table}

\paragraph{Skill transfer.}
\label{sec:skill-transfer}

To test whether learned skills remain effective beyond the model used during optimization, we transfer the benchmark-specific skills learned with Qwen3.8-27B as the target and GPT-5.5 as the optimizer to Qwen3.6-35B-A3B and Qwen3.5-4B. Table~\ref{tab:skill-transfer} compares skills optimized for each receiving model (Origin) with the transferred skills, which are used without further context optimization. Transfer yields a slightly higher overall score on both receiving models and matches the original skill on ALFWorld. The gains are not uniform: OfficeQA decreases by 1.1 percentage points on Qwen3.6-35B-A3B, and LiveMath decreases by 0.8 points on Qwen3.5-4B. These results suggest that the learned guidance can be applied across the tested Qwen models while retaining its overall effectiveness without additional context optimization.

\input{sections/optimizer_results_table}

\paragraph{Optimizer choice.}
\label{sec:optimizer-choice}

To examine whether the observed gains depend on using GPT-5.5 as the optimizer, we also optimize skills with Qwen3.8-27B while keeping the target fixed as Qwen3.8-27B. Table~\ref{tab:optimizer-choice} reports an overall score of 69.7\% with the Qwen optimizer, compared with 71.0\% with GPT-5.5. GPT-5.5 produces higher scores on all six benchmarks, with an overall advantage of 1.3 percentage points. Nevertheless, the Qwen optimizer improves over the Base Model in Table~\ref{tab:main-results} on every benchmark, with an average gain of 14.1 points. TextualRL therefore remains effective when the same model serves as both target and optimizer in this setting, although the optimizer choice affects the final task performance.

\begin{figure}[!t]
    \centering
    \includegraphics[width=\linewidth]{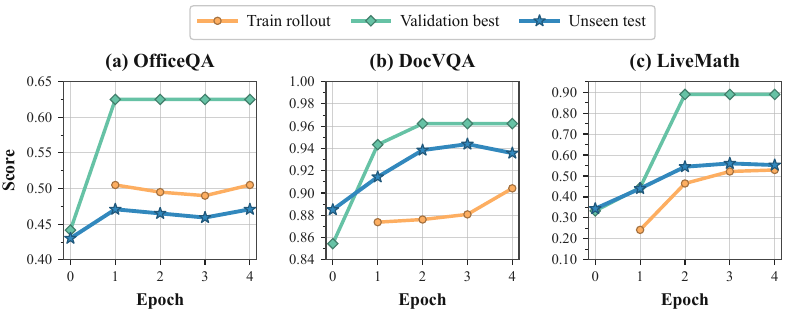}
    \caption{Epoch curves for TextualRL on Qwen3.8-27B.}
    \label{fig:epoch-main}
\end{figure}

\paragraph{Training dynamics.}
To examine when context optimization yields gains and whether additional epochs continue to improve unseen performance, we track training, validation, and test scores across four epochs on Qwen3.8-27B. Figure~\ref{fig:epoch-main} shows that most test improvement occurs early, with different gains across benchmarks. By epoch 2, DocVQA improves from 88.5\% to 93.9\%, and LiveMath from 34.4\% to 54.4\%. Their validation scores then plateau, while test scores fluctuate within 93.6--94.4\% and 54.4--56.0\%, respectively. OfficeQA reaches 47.1\% on test after the first epoch and shows no sustained improvement thereafter, despite a validation score of 62.5\%. Thus, further optimization does not consistently improve unseen performance. On OfficeQA and LiveMath, validation scores improve much more than test scores. This highlights a generalization challenge: instructions favored by a fixed validation set may be less effective on unseen tasks.

%% file: sections/main_results_table.tex
\begin{table}[!t]
\centering
\small
\setlength{\tabcolsep}{5.75pt}
\renewcommand{\arraystretch}{1.0}
\caption{Main test results (\%). Bold and underlining mark the highest and second-highest distinct scores in each column within each target model. Subscripts on TextualRL scores show percentage-point gains over the corresponding Base Model.}
\label{tab:main-results}
\begin{tabular}{@{}lccccccc}
\toprule
Model & Skill Source & SearchQA & Spreadsheet & OfficeQA & DocVQA & LiveMath & ALFWorld \\
\midrule
\multirow{8}{*}{\textbf{GPT-5.5}} & Base Model & 77.7 & 41.8 & 33.1 & 78.8 & 37.6 & 83.6 \\
 & Human skill & 81.8 & 72.9 & 66.9 & 90.1 & 38.4 & 91.8 \\
 & LLM skill & 80.9 & 43.2 & 51.7 & 89.6 & 40.0 & \underline{93.3} \\
 & Trace2Skill & 82.4 & 49.6 & 65.7 & 90.6 & 52.0 & 87.3 \\
 & TextGrad & 81.4 & 41.1 & 42.0 & 87.2 & 49.2 & 82.8 \\
 & GEPA & 84.8 & 73.6 & 63.9 & 89.1 & 43.2 & 85.8 \\
 & SkillOpt & \underline{87.3} & \underline{80.7} & \underline{72.1} & \underline{91.2} & \underline{66.9} & \textbf{95.5} \\
\rowcolor{textualrlblue}
\cellcolor{white} & \textbf{TextualRL} & \textbf{88.9}\rlap{\hspace{0.15em}\textsubscript{\textcolor{green!40!black}{+11.2}}} & \textbf{82.5}\rlap{\hspace{0.15em}\textsubscript{\textcolor{green!40!black}{+40.7}}} & \textbf{72.7}\rlap{\hspace{0.15em}\textsubscript{\textcolor{green!40!black}{+39.6}}} & \textbf{92.5}\rlap{\hspace{0.15em}\textsubscript{\textcolor{green!40!black}{+13.7}}} & \textbf{67.2}\rlap{\hspace{0.15em}\textsubscript{\textcolor{green!40!black}{+29.6}}} & \textbf{95.5}\rlap{\hspace{0.15em}\textsubscript{\textcolor{green!40!black}{+11.9}}} \\
\midrule
\multirow{8}{*}{\textbf{GPT-5.4}} & Base Model & 76.9 & 41.4 & 50.0 & 77.6 & 36.8 & 75.4 \\
 & Human skill & 80.1 & 59.3 & 59.3 & 88.5 & 41.6 & 74.6 \\
 & LLM skill & 79.4 & 46.1 & 20.4 & 90.4 & 36.8 & 83.6 \\
 & Trace2Skill & 81.2 & 53.2 & 51.2 & 89.3 & 40.0 & 73.1 \\
 & TextGrad & 82.5 & 38.6 & 54.7 & 85.3 & 36.6 & 78.4 \\
 & GEPA & 82.4 & 61.1 & 60.4 & 88.3 & 41.6 & 74.6 \\
 & SkillOpt & \underline{83.1} & \underline{62.5} & \underline{62.8} & \underline{91.2} & \underline{44.0} & \underline{91.0} \\
\rowcolor{textualrlblue}
\cellcolor{white} & \textbf{TextualRL} & \textbf{85.5}\rlap{\hspace{0.15em}\textsubscript{\textcolor{green!40!black}{+8.6}}} & \textbf{63.6}\rlap{\hspace{0.15em}\textsubscript{\textcolor{green!40!black}{+22.2}}} & \textbf{64.5}\rlap{\hspace{0.15em}\textsubscript{\textcolor{green!40!black}{+14.5}}} & \textbf{91.7}\rlap{\hspace{0.15em}\textsubscript{\textcolor{green!40!black}{+14.1}}} & \textbf{47.2}\rlap{\hspace{0.15em}\textsubscript{\textcolor{green!40!black}{+10.4}}} & \textbf{91.8}\rlap{\hspace{0.15em}\textsubscript{\textcolor{green!40!black}{+16.4}}} \\
\midrule
\multirow{8}{*}{\shortstack[l]{\textbf{Qwen3.5-}\\\textbf{4B}}} & Base Model & 68.1 & 9.3 & 14.5 & 86.9 & 22.4 & 30.6 \\
 & Human skill & 66.3 & 16.4 & 22.7 & 87.8 & 18.4 & 28.4 \\
 & LLM skill & 65.0 & 10.4 & 20.9 & 88.0 & 28.8 & 55.2 \\
 & Trace2Skill & 68.5 & 19.3 & 16.3 & 88.0 & 27.2 & 64.9 \\
 & TextGrad & 60.7 & 13.9 & 20.9 & 85.6 & 10.6 & 53.7 \\
 & GEPA & 68.6 & 16.9 & 22.7 & 85.1 & 28.8 & 60.4 \\
 & SkillOpt & \underline{71.2} & \underline{23.9} & \underline{29.7} & \underline{89.0} & \underline{52.0} & \underline{81.3} \\
\rowcolor{textualrlblue}
\cellcolor{white} & \textbf{TextualRL} & \textbf{73.9}\rlap{\hspace{0.15em}\textsubscript{\textcolor{green!40!black}{+5.8}}} & \textbf{27.1}\rlap{\hspace{0.15em}\textsubscript{\textcolor{green!40!black}{+17.8}}} & \textbf{32.0}\rlap{\hspace{0.15em}\textsubscript{\textcolor{green!40!black}{+17.5}}} & \textbf{89.3}\rlap{\hspace{0.15em}\textsubscript{\textcolor{green!40!black}{+2.4}}} & \textbf{56.0}\rlap{\hspace{0.15em}\textsubscript{\textcolor{green!40!black}{+33.6}}} & \textbf{82.1}\rlap{\hspace{0.15em}\textsubscript{\textcolor{green!40!black}{+51.5}}} \\
\midrule
\multirow{8}{*}{\shortstack[l]{\textbf{Qwen3.6-}\\\textbf{35B-A3B}}} & Base Model & 72.7 & 38.2 & 45.9 & 87.6 & 31.2 & 59.7 \\
 & Human skill & 74.1 & 44.3 & 41.9 & 88.4 & 29.6 & 44.8 \\
 & LLM skill & 72.6 & 42.9 & 46.5 & 88.4 & 24.8 & 60.4 \\
 & Trace2Skill & 75.4 & 33.2 & 32.0 & 90.4 & 29.6 & 70.9 \\
 & TextGrad & 76.4 & 22.9 & 33.7 & 84.5 & 7.2 & 67.9 \\
 & GEPA & 75.8 & 45.4 & 43.6 & 88.0 & 31.2 & 73.9 \\
 & SkillOpt & \underline{80.3} & \underline{47.5} & \underline{47.1} & \underline{91.4} & \underline{41.6} & \underline{82.1} \\
\rowcolor{textualrlblue}
\cellcolor{white} & \textbf{TextualRL} & \textbf{81.2}\rlap{\hspace{0.15em}\textsubscript{\textcolor{green!40!black}{+8.5}}} & \textbf{52.1}\rlap{\hspace{0.15em}\textsubscript{\textcolor{green!40!black}{+13.9}}} & \textbf{49.4}\rlap{\hspace{0.15em}\textsubscript{\textcolor{green!40!black}{+3.5}}} & \textbf{92.8}\rlap{\hspace{0.15em}\textsubscript{\textcolor{green!40!black}{+5.2}}} & \textbf{44.8}\rlap{\hspace{0.15em}\textsubscript{\textcolor{green!40!black}{+13.6}}} & \textbf{83.6}\rlap{\hspace{0.15em}\textsubscript{\textcolor{green!40!black}{+23.9}}} \\
\midrule
\multirow{8}{*}{\shortstack[l]{\textbf{Qwen3.8-}\\\textbf{27B}}} & Base Model & 73.6 & 43.6 & 43.0 & 88.5 & 34.4 & 50.8 \\
 & Human skill & 74.9 & 48.6 & 43.6 & 89.0 & 32.8 & 47.8 \\
 & LLM skill & 74.5 & 46.4 & 42.4 & 88.8 & 32.0 & 59.7 \\
 & Trace2Skill & 76.9 & 45.0 & 43.0 & 90.4 & 37.6 & 72.4 \\
 & TextGrad & 77.6 & 39.3 & 36.1 & 88.2 & 35.2 & 56.7 \\
 & GEPA & 78.3 & 50.7 & 44.2 & 89.0 & 40.0 & 75.4 \\
 & SkillOpt & \underline{80.4} & \underline{56.4} & \underline{46.5} & \underline{93.3} & \underline{48.0} & \underline{82.1} \\
\rowcolor{textualrlblue}
\cellcolor{white} & \textbf{TextualRL} & \textbf{82.7}\rlap{\hspace{0.15em}\textsubscript{\textcolor{green!40!black}{+9.1}}} & \textbf{60.7}\rlap{\hspace{0.15em}\textsubscript{\textcolor{green!40!black}{+17.1}}} & \textbf{47.7}\rlap{\hspace{0.15em}\textsubscript{\textcolor{green!40!black}{+4.7}}} & \textbf{94.4}\rlap{\hspace{0.15em}\textsubscript{\textcolor{green!40!black}{+5.9}}} & \textbf{56.0}\rlap{\hspace{0.15em}\textsubscript{\textcolor{green!40!black}{+21.6}}} & \textbf{84.3}\rlap{\hspace{0.15em}\textsubscript{\textcolor{green!40!black}{+33.5}}} \\
\bottomrule
\end{tabular}
\end{table}

%% file: sections/ablation_results_table.tex
\begin{table}[!t]
\centering
\small
\setlength{\tabcolsep}{3pt}
\renewcommand{\arraystretch}{1.0}
\caption{Ablations on Qwen3.8-27B (\%). Full TextualRL is the shared reference for both groups.}
\label{tab:ablations}
\begin{tabular}{lcccccc}
\toprule
Setting & SearchQA & Spreadsheet & OfficeQA & DocVQA & LiveMath & ALFWorld \\
\midrule
\rowcolor{textualrlblue}
\textbf{TextualRL (full)} & \textbf{82.7} & \textbf{60.7} & \textbf{47.7} & \textbf{94.4} & \textbf{56.0} & \textbf{84.3} \\
\midrule
\multicolumn{7}{l}{\textit{Groupwise Policy Critique}} \\
w/o outcome routing & \underline{81.8} & \underline{58.9} & \underline{46.5} & 92.5 & \underline{53.6} & \underline{83.6} \\
Single rollout ($K=1$) & 80.7 & 57.5 & \underline{46.5} & \underline{93.3} & 50.4 & 82.1 \\
\midrule
\multicolumn{7}{l}{\textit{Cross-Group Policy Update}} \\
Intra-group cards only & \underline{82.1} & \underline{59.3} & \underline{47.1} & 93.0 & 53.6 & \underline{83.6} \\
Remove cards without edits & 80.9 & 58.2 & 45.9 & \underline{93.6} & \underline{54.4} & 82.8 \\
w/o proposal review & 81.4 & 57.1 & 45.4 & 92.8 & 50.4 & 82.8 \\
\bottomrule
\end{tabular}
\end{table}

%% file: sections/run_stability_table.tex
\begin{table}[!t]
\centering
\small
\setlength{\tabcolsep}{2pt}
\renewcommand{\arraystretch}{1.05}
\caption{Mean $\pm$ sample standard deviation over three independent runs on Qwen3.8-27B (\%).}
\label{tab:run-stability}
\begin{tabular}{lccccccc}
\toprule
Method & SearchQA & Spreadsheet & OfficeQA & DocVQA & LiveMath & ALFWorld & Overall \\
\midrule
SkillOpt & $79.93\!\pm\!0.47$ & $56.55\!\pm\!0.54$ & $45.93\!\pm\!0.59$ & $92.87\!\pm\!0.41$ & $47.73\!\pm\!1.22$ & $81.59\!\pm\!0.86$ & $67.43\!\pm\!0.33$ \\
\rowcolor{textualrlblue}
\textbf{TextualRL} & $\mathbf{82.62}\!\pm\!0.16$ & $\mathbf{60.24}\!\pm\!0.55$ & $\mathbf{47.09}\!\pm\!0.58$ & $\mathbf{94.03}\!\pm\!0.41$ & $\mathbf{56.00}\!\pm\!0.80$ & $\mathbf{84.08}\!\pm\!1.14$ & $\mathbf{70.68}\!\pm\!0.26$ \\
\bottomrule
\end{tabular}
\vspace{-2mm}
\end{table}

%% file: sections/rollout_count_table.tex
\begin{table}[!t]
\centering
\small
\setlength{\tabcolsep}{4.5pt}
\renewcommand{\arraystretch}{1.05}
\caption{Effect of training rollout count $K$ on Qwen3.8-27B (\%).}
\label{tab:rollout-count}
\begin{tabular}{lcccccccc}
\toprule
Method & $K$ & SearchQA & Spreadsheet & OfficeQA & DocVQA & LiveMath & ALFWorld & Overall \\
\midrule
\multirow{4}{*}{\textbf{TextualRL}} & 1 & 80.7 & 57.5 & 46.5 & 93.3 & 50.4 & 82.1 & 68.4 \\
 & 2 & 81.4 & 58.9 & \underline{47.7} & 93.6 & \underline{53.6} & 82.8 & 69.7 \\
 & 4 & \underline{82.7} & \textbf{60.7} & \underline{47.7} & \textbf{94.4} & \textbf{56.0} & \underline{84.3} & \underline{71.0} \\
 & 8 & \textbf{82.9} & \underline{60.0} & \textbf{48.8} & \underline{94.1} & \textbf{56.0} & \textbf{85.1} & \textbf{71.1} \\
\bottomrule
\end{tabular}
\vspace{-2mm}
\end{table}

%% file: sections/harness_results_table.tex
\begin{table}[!t]
\centering
\small
\setlength{\tabcolsep}{5.75pt}
\renewcommand{\arraystretch}{1.05}
\caption{Test results (\%) with GPT-5.5 under different agent harnesses.}
\label{tab:harness-results}
\begin{tabular}{@{}lcccccc@{\hspace{16pt}}}
\toprule
Harness & Skill Source & SearchQA & Spreadsheet & OfficeQA & DocVQA & LiveMath \\
\midrule
\multirow{6}{*}{\textbf{Codex}} & Base Model & 81.8 & 27.5 & 38.3 & 87.2 & 35.2 \\
 & Human skill & 84.1 & 50.7 & 40.0 & 88.8 & 48.8 \\
 & LLM skill & 83.4 & 25.0 & 34.3 & 89.8 & 45.6 \\
 & EvoSkill & 61.4 & 67.5 & 42.4 & 89.3 & 63.2 \\
 & SkillOpt & \underline{87.3} & \underline{85.0} & \underline{51.1} & \underline{92.2} & \textbf{78.4} \\
\rowcolor{textualrlblue}
\cellcolor{white} & \textbf{TextualRL} & \textbf{88.6}\rlap{\hspace{0.15em}\textsubscript{\textcolor{green!40!black}{+6.8}}} & \textbf{85.7}\rlap{\hspace{0.15em}\textsubscript{\textcolor{green!40!black}{+58.2}}} & \textbf{56.4}\rlap{\hspace{0.15em}\textsubscript{\textcolor{green!40!black}{+18.1}}} & \textbf{92.8}\rlap{\hspace{0.15em}\textsubscript{\textcolor{green!40!black}{+5.6}}} & \underline{76.8}\rlap{\hspace{0.15em}\textsubscript{\textcolor{green!40!black}{+41.6}}} \\
\midrule
\multirow{6}{*}{\shortstack[l]{\textbf{Claude}\\\textbf{Code}}} & Base Model & 81.9 & 22.1 & 57.6 & 86.6 & 40.8 \\
 & Human skill & 83.7 & 37.1 & 66.3 & 88.0 & 44.0 \\
 & LLM skill & 82.4 & 37.1 & 56.4 & 89.6 & 44.8 \\
 & EvoSkill & 84.0 & 75.0 & 70.3 & 87.2 & 52.0 \\
 & SkillOpt & \underline{85.9} & \underline{80.4} & \underline{71.5} & \underline{90.1} & \underline{56.5} \\
\rowcolor{textualrlblue}
\cellcolor{white} & \textbf{TextualRL} & \textbf{87.1}\rlap{\hspace{0.15em}\textsubscript{\textcolor{green!40!black}{+5.2}}} & \textbf{83.9}\rlap{\hspace{0.15em}\textsubscript{\textcolor{green!40!black}{+61.8}}} & \textbf{73.3}\rlap{\hspace{0.15em}\textsubscript{\textcolor{green!40!black}{+15.7}}} & \textbf{90.9}\rlap{\hspace{0.15em}\textsubscript{\textcolor{green!40!black}{+4.3}}} & \textbf{57.6}\rlap{\hspace{0.15em}\textsubscript{\textcolor{green!40!black}{+16.8}}} \\
\bottomrule
\end{tabular}
\end{table}

%% file: sections/skill_transfer_table.tex
\begin{table}[!t]
\centering
\small
\setlength{\tabcolsep}{3pt}
\renewcommand{\arraystretch}{1.05}
\caption{Skill transfer from Qwen3.8-27B to two target models (\%).}
\label{tab:skill-transfer}
\begin{tabular}{@{}lcccccccc@{}}
\toprule
Target model & Skill & SearchQA & Spreadsheet & OfficeQA & DocVQA & LiveMath & ALFWorld & Overall \\
\midrule
\multirow{2}{*}{\shortstack[l]{Qwen3.6-\\35B-A3B}} & Origin & 81.2 & 52.1 & 49.4 & 92.8 & 44.8 & 83.6 & 67.3 \\
 & Transferred & 81.4 & 53.2 & 48.3 & 93.3 & 47.2 & 83.6 & 67.8 \\
\midrule
\multirow{2}{*}{Qwen3.5-4B} & Origin & 73.9 & 27.1 & 32.0 & 89.3 & 56.0 & 82.1 & 60.1 \\
 & Transferred & 74.3 & 27.9 & 32.0 & 90.1 & 55.2 & 82.1 & 60.3 \\
\bottomrule
\end{tabular}
\end{table}

%% file: sections/optimizer_results_table.tex
\begin{table}[!t]
\centering
\small
\setlength{\tabcolsep}{4pt}
\renewcommand{\arraystretch}{1.05}
\caption{Optimizer comparison with Qwen3.8-27B as the fixed target (\%).}
\label{tab:optimizer-choice}
\begin{tabular}{@{}lccccccc@{}}
\toprule
Optimizer & SearchQA & Spreadsheet & OfficeQA & DocVQA & LiveMath & ALFWorld & Overall \\
\midrule
GPT-5.5 & 82.7 & 60.7 & 47.7 & 94.4 & 56.0 & 84.3 & 71.0 \\
Qwen3.8-27B & 81.5 & 58.6 & 46.5 & 93.9 & 54.4 & 83.6 & 69.7 \\
\bottomrule
\end{tabular}
\end{table}

%% file: sections/conclusion.tex
\section{Conclusion}
\label{sec:conclusion}

We introduce TextualRL, a gradient-free reinforcement learning framework for optimizing the textual context of a frozen LLM agent through task rewards. Groupwise Policy Critique uses same-task rollout comparisons and cross-task analysis to propose context edits according to the available outcome patterns. Cross-Group Policy Update reviews these proposals against pooled trajectory evidence, including observations that produce no edit, to refine their content and applicability before validation-based selection. Across five target models and six benchmarks, TextualRL achieves the highest six-benchmark average for every target. Ablation studies and additional experiments support the effectiveness of both modules and the consistency of the observed improvements.

%% file: sections/appendix.tex
\section{Method Implementation Details}
\label{app:method-details}

\begin{algorithm}[!t]
\caption{A step of TextualRL with fixed target parameters}
\label{alg:textualrl-step}
\begin{algorithmic}[1]
\Require Context $C$, recorded validation score $s$, training batch $B$, trajectories per task $K$, validation set $\mathcal V$, edit budget $L$, shared history $H$
\State For each $x\in B$, collect $D_x$ under $C$ and compute $\mathbf r_x$ (Equation~\ref{eq:task-block})
\State Route each $D_x$ using $\mathbf r_x$, then consolidate duplicate trajectories within the task
\State Form groups $\mathcal G$: one per mixed task and multiple tasks per homogeneous group
\State In parallel, obtain $(P_g,E_g)$ for each $g\in\mathcal G$ from $\{(x,D_x,\mathbf r_x):x\in g\}$, $C$, and $H$
\State Pool proposals $P$ and valid current-step cards $E$ independently
\State Consolidate $P$ hierarchically and review the candidates against $E$ to obtain $\widetilde P$
\State Select at most $L$ edits from $\widetilde P$ and filter dependencies on unavailable runtime information
\If{no edits remain}
    \State \Return $C,s$ \Comment{No candidate update}
\EndIf
\State Assemble $\Delta C$ and apply it to obtain $C'\gets C+\Delta C$
\State Evaluate $v'\gets\widehat V(C')$ on $\mathcal V$
\If{$v'>s$}
    \State Adopt $C',v'$ and update the best validation-scored snapshot when improved
\Else
    \State Retain $C,s$ and record the rejected edit combination and its score
\EndIf
\State Add rollout and rejection feedback to $H$ for subsequent proposals
\end{algorithmic}
\end{algorithm}

\paragraph{Proposal consolidation.}
Mixed-task and all-failure proposals enter the corrective branch, while all-success proposals enter the preservation branch. The optimizer recursively merges proposal batches within each branch before combining the two branches. The final merge prioritizes corrective edits unless they conflict with a supported success pattern.

\paragraph{Edit application.}
Patch operations are applied deterministically while preserving protected scaffold regions. The per-edit dependency filter uses explicit patterns with limited negation handling to detect dependencies on information unavailable during target execution, such as gold answers, grader feedback, or validation scores. It removes an affected edit as a whole and retains other valid edits from the same proposal.

\paragraph{Evidence provenance.}
We retain evidence cards whose task and rollout references resolve to the displayed trajectories and whose supporting quotes occur in the referenced trace. Repeated cards for the same task and trajectory are consolidated. References attached to reviewed edits are checked against the pooled cards. If no valid cards remain, the optimizer uses ordinary proposal merging.

\paragraph{Repeated proposals.}
The step buffer screens exact or near-equivalent repetitions of a rejected edit combination at the combination level. An individual edit family is blocked only after it appears in at least three distinct rejected candidates and no accepted candidate.

\section{Implementation and Data Notes}
\label{app:implementation}

\subsection{Data Splits and Metrics}
\label{app:benchmarks}

Table~\ref{tab:splits} summarizes the data splits and evaluation metrics for all six benchmarks, with counts referring to the experimental subsets. We report binary task accuracy or success rate in percentages. SpreadsheetBench requires all test cases for a task to pass. DocVQA uses hard accuracy, with ANLS $\geq 0.999$ counted as correct. Overall performance is the unweighted mean across the six benchmarks.

For ALFWorld, we use the official SkillOpt release with 39 training, 18 validation, and 134 test environments. The released validation manifest selects 18 environments from the full \texttt{valid\_seen} pool of 140, whereas the SkillOpt paper~\citep{yang2026skillopt} describes 140 validation environments. Our update acceptance and checkpoint selection use the 18 environments in the released manifest.

\begin{table}[!t]
\centering
\normalsize
\setlength{\tabcolsep}{8pt}
\renewcommand{\arraystretch}{1.12}
\caption{Benchmark splits and evaluation metrics.}
\label{tab:splits}
\begin{tabular}{lcccl}
\toprule
Benchmark & Train & Validation & Test & Metric \\
\midrule
SearchQA & 400 & 200 & 1,400 & Normalized exact match \\
SpreadsheetBench & 80 & 40 & 280 & All-case success \\
OfficeQA & 50 & 24 & 172 & Normalized exact match \\
DocVQA & 107 & 53 & 374 & Hard accuracy \\
LiveMath & 35 & 17 & 125 & Choice accuracy \\
ALFWorld & 39 & 18 & 134 & Task success \\
\bottomrule
\end{tabular}
\end{table}

\subsection{Critique and Merge Batches}
\label{app:critique-batches}

In full TextualRL, the critique minibatch size of eight counts distinct task blocks, whereas the merge minibatch size counts proposal bundles. Each mixed task is analyzed separately with all of its unique sibling trajectories. Within each homogeneous route, task blocks are grouped into batches of eight. A final batch containing two to seven tasks is processed as is. If the final batch contains only one task, it is combined with the preceding batch, producing a batch of nine. A homogeneous route containing only one task is skipped.

Outcome routing uses the original $K$ rollout outcomes before deduplication. Duplicate trajectories are consolidated only within the same task and outcome label, using the recorded conversation with whitespace normalized and timing and request metadata removed. Each retained trajectory keeps its multiplicity and original rollout references. Deduplication reduces the trajectories shown to the critic but does not change $K$, the task's outcome category, or the number of distinct tasks supporting a proposal.

Hierarchical merging processes up to eight proposal bundles per call, and each bundle may contain multiple edits. A final batch of two to seven bundles is merged normally. A singleton bundle passes directly to the next level without a merge call. These batch sizes are separate from the edit budget, which bounds the number of edits applied to form the candidate context.

For \emph{w/o outcome routing}, we pool the step's trajectories and form analyst minibatches of eight trajectories. We adapt the mixed-outcome analyst prompt by removing the requirements that trajectories belong to a single task and include both successes and failures. The resulting unified prompt asks the analyst to produce edit proposals and evidence cards from the available trajectories and outcomes.

In \emph{intra-group cards only}, each critique group $g\in\mathcal{G}_t$ corresponds to one analyst call. Separate calls within the same outcome route remain distinct groups. The ablation retains hierarchical consolidation followed by evidence review during the final merge. For each merged edit, review is restricted to the union of evidence cards from the critique groups that contributed its source proposals. For example, an edit formed from proposals by groups $A$ and $B$ can use $E_A\cup E_B$, while cards from other groups are excluded.

\subsection{Generation and Evaluation}
\label{app:generation-protocol}

Training rollouts use a temperature of 0.7, while validation and test inference use a temperature of 0. Top-p is left at the serving backend's default. The random seed is set to 42. GPT-5.5 and GPT-5.4 use \texttt{medium} reasoning effort.

For each target--benchmark pair, context optimization collects $K=4$ rollouts per training task under the same current context, except in the rollout-count ablation. Validation selects the context used for test evaluation. The TextualRL scores in Table~\ref{tab:main-results} are from individual optimization runs. The run-stability experiment separately repeats the complete optimization and validation-based checkpoint selection three times for TextualRL and SkillOpt on Qwen3.8-27B, with fixed data splits. Its reported variation is across independently optimized contexts, rather than repeated evaluations of a single context.

Each rollout starts with a fresh interaction history and the same task inputs, with separate sample identifiers for its saved outputs. Spreadsheet rollouts operate on independent copies of the input workbook. OfficeQA rollouts start new conversations over the supplied documents. ALFWorld rollouts initialize and reset separate environments using the task's game file, with environment seeds derived from the batch seed and environment index.

\subsection{Checkpoint Selection}
For each run, we select the context with the highest recorded validation score for test evaluation. We track this checkpoint separately from the active context after the final update. Checkpoint selection uses validation scores exclusively, and the selected context is then supplied to the frozen LLM agent for test tasks.

\subsection{Token Accounting}
\label{app:token-usage}

\begin{table}[!t]
\centering
\small
\setlength{\tabcolsep}{4pt}
\renewcommand{\arraystretch}{1.05}
\caption{Token usage with Qwen3.8-27B as the target model with GPT-5.5 as optimizer. Iteration usage includes candidate validation. Inference usage sums input and visible output text per task.}
\label{tab:token-usage}
\begin{tabular}{lcccccc}
\toprule
\multirow{2}{*}{Benchmark} & \multirow{2}{*}{Method} & \multirow{2}{*}{\shortstack{Final\\tokens}} & \multicolumn{3}{c}{Recorded iteration tokens (M)} & \multirow{2}{*}{\shortstack{Inference text\\tokens / task}} \\
\cmidrule(lr){4-6}
 & & & Target & Optimizer & Total & \\
\midrule
\multirow{2}{*}{SearchQA} & SkillOpt & 937 & 31.33 & 7.67 & 39.00 & 2,698.2 \\
 & TextualRL & 2,422 & 66.26 & 13.57 & 79.83 & 5,035.4 \\
\addlinespace[2pt]
\multirow{2}{*}{Spreadsheet} & SkillOpt & 2,908 & 49.62 & 4.99 & 54.60 & 9,044.2 \\
 & TextualRL & 1,397 & 32.99 & 8.35 & 41.34 & 4,145.6 \\
\addlinespace[2pt]
\multirow{2}{*}{OfficeQA} & SkillOpt & 1,589 & 44.18 & 4.70 & 48.88 & 105,784.5 \\
 & TextualRL & 974 & 142.87 & 7.21 & 150.08 & 137,714.1 \\
\addlinespace[2pt]
\multirow{2}{*}{DocVQA} & SkillOpt & 415 & 4.60 & 0.63 & 5.22 & 741.0 \\
 & TextualRL & 659 & 10.76 & 0.79 & 11.56 & 1,132.2 \\
\addlinespace[2pt]
\multirow{2}{*}{LiveMath} & SkillOpt & 1,534 & 1.59 & 1.29 & 2.88 & 7,889.8 \\
 & TextualRL & 1,913 & 4.98 & 4.34 & 9.32 & 8,410.2 \\
\addlinespace[2pt]
\multirow{2}{*}{ALFWorld} & SkillOpt & 516 & 8.55 & 0.28 & 8.83 & 24,929.9 \\
 & TextualRL & 1,600 & 25.21 & 0.83 & 26.05 & 43,786.2 \\
\bottomrule
\end{tabular}
\end{table}

The settings and selected skills for Table~\ref{tab:token-usage} follow the corresponding Qwen3.8-27B experiments in Table~\ref{tab:main-results}, with GPT-5.5 as optimizer. Final refers to the skill selected for inference. Table~\ref{tab:token-usage} reports recorded iteration subtotals. Target counts sum input and output tokens for training rollouts and candidate validation within saved steps. Optimizer counts cover optimization calls in those same steps. Totals include all saved steps, including those after the selected checkpoint. The accounting scope excludes subsequent unseen evaluation, initial validation outside these steps, and unrecorded recovery or retry usage. M denotes one million tokens. Totals are calculated from the original integer counts, and each displayed column is rounded independently.

Inference counts use the Qwen3.8-27B tokenizer with \texttt{add\_special\_tokens=False}. Input message text is accumulated across all saved decisions for each task, with repeated history counted at each decision. Output counts include the saved visible model text. Their sum is calculated before rounding and averaged over tasks with saved trajectories. Each token contributes equally to this measure of recorded text volume, independent of token prices, caching, or computational cost.

TextualRL records 2.05--3.24 times SkillOpt's iteration tokens on five benchmarks. Collecting $K=4$ trajectories per task incurs sampling cost to obtain comparative feedback, analogous to group-based RL such as GRPO~\citep{shao2024deepseekmath}. Trajectory analysis, evidence review, and candidate validation also consume optimization tokens. Review reuses existing rollouts. At inference, these optimization procedures are absent, and token usage depends on the learned context and execution history. The corresponding increases are smaller, at 6.6--86.6\%. Spreadsheet instead uses 24.3\% fewer recorded iteration tokens and 54.2\% fewer inference text tokens.

\subsection{Evidence Review Audit}
\label{app:evidence-review-audit}

To examine how review uses recorded evidence, we audit runs with Qwen3.8-27B as target and GPT-5.5 as optimizer across all six benchmarks. We inspect all 76 saved steps, containing 464 merged edits and 578 raw evidence references. We randomly sample up to 20 reviewed edits per benchmark with seed 42, including all 13 available LiveMath edits, yielding 113 edits and 153 references. Two separate GPT-6 Astra sessions independently assess the same sample using an identical rubric, without seeing the original review labels, validation outcomes, or each other's judgments. Each assesses card faithfulness to the available runtime text, the card's relation to the final reviewed edit, and the grounding of the edit's scope.

\begin{table}[!t]
\centering
\small
\setlength{\tabcolsep}{5pt}
\renewcommand{\arraystretch}{1.15}
\caption{Evidence review audit on Qwen3.8-27B runs.}
\label{tab:evidence-review-audit}
\begin{tabular}{@{}p{0.26\linewidth}p{0.26\linewidth}p{0.42\linewidth}@{}}
\toprule
Audit item & Result & Measurement \\
\midrule
Reference traceability & 99.13\% (573/578) & Card resolves and its quote matches the source trajectory. \\
LLM relation agreement & 94.77\% (145/153) & Agreement between two GPT-6 Astra assessments, $\kappa=0.715$. \\
Reported review changes & 8.48\% (37/436) & Optimizer reports \texttt{narrow} or \texttt{revise}. \\
\bottomrule
\end{tabular}
\end{table}

Table~\ref{tab:evidence-review-audit} separates source traceability from model assessment. References are checked against the current-step card pool after the existing card validation procedure, and all raw review references remain in the denominator. Relation labels are \emph{supports}, \emph{limits}, \emph{contradicts}, \emph{irrelevant}, and \emph{unclear}, always evaluated against the final reviewed edit. The two assessments label 134/153 card statements as supporting the rule. Agreement measures consistency between two evaluations by the same model, rather than correctness against human judgments.

These findings support the auditability of Cross-Group Policy Update. Traceable references allow readers to inspect the observations cited for individual rules, while the two assessments agree on a supporting relation for most sampled references (87.58\%). The review therefore exposes links between evidence and edits that can be examined beyond the final task score. The optimizer reports narrowing or revision for 37 of the 436 edits with review metadata. This pattern is consistent with review serving as a selective refinement stage, where proposals can be retained after evidence comparison and changes can target their content or applicability.

\paragraph{Effect of reviewed updates.}
To examine whether evidence review improves task performance beyond proposal consolidation, we evaluate three context variants for one selected update each on SearchQA, OfficeQA, and DocVQA, using Qwen3.8-27B as the target and GPT-5.5 as optimizer. \emph{Old skill} is the context before the selected update. \emph{New skill w/o review} applies the consolidated proposal without evidence review, while \emph{New skill} applies the corresponding reviewed edit. Each comparison uses the same historical context and unseen-test split.

\begin{table}[!t]
\centering
\small
\setlength{\tabcolsep}{10pt}
\renewcommand{\arraystretch}{1.1}
\caption{Unseen-test performance (\%) of three context variants for selected updates.}
\label{tab:review-update-results}
\begin{tabular}{@{}lccc@{}}
\toprule
Benchmark & Old skill & New skill w/o review & New skill \\
\midrule
SearchQA & 81.57 & 82.07 & \textbf{82.71} \\
OfficeQA & 45.35 & 45.93 & \textbf{47.09} \\
DocVQA & 89.57 & 90.37 & \textbf{91.44} \\
\bottomrule
\end{tabular}
\end{table}

\begin{figure}[!t]
    \centering
    \includegraphics[width=0.9\linewidth]{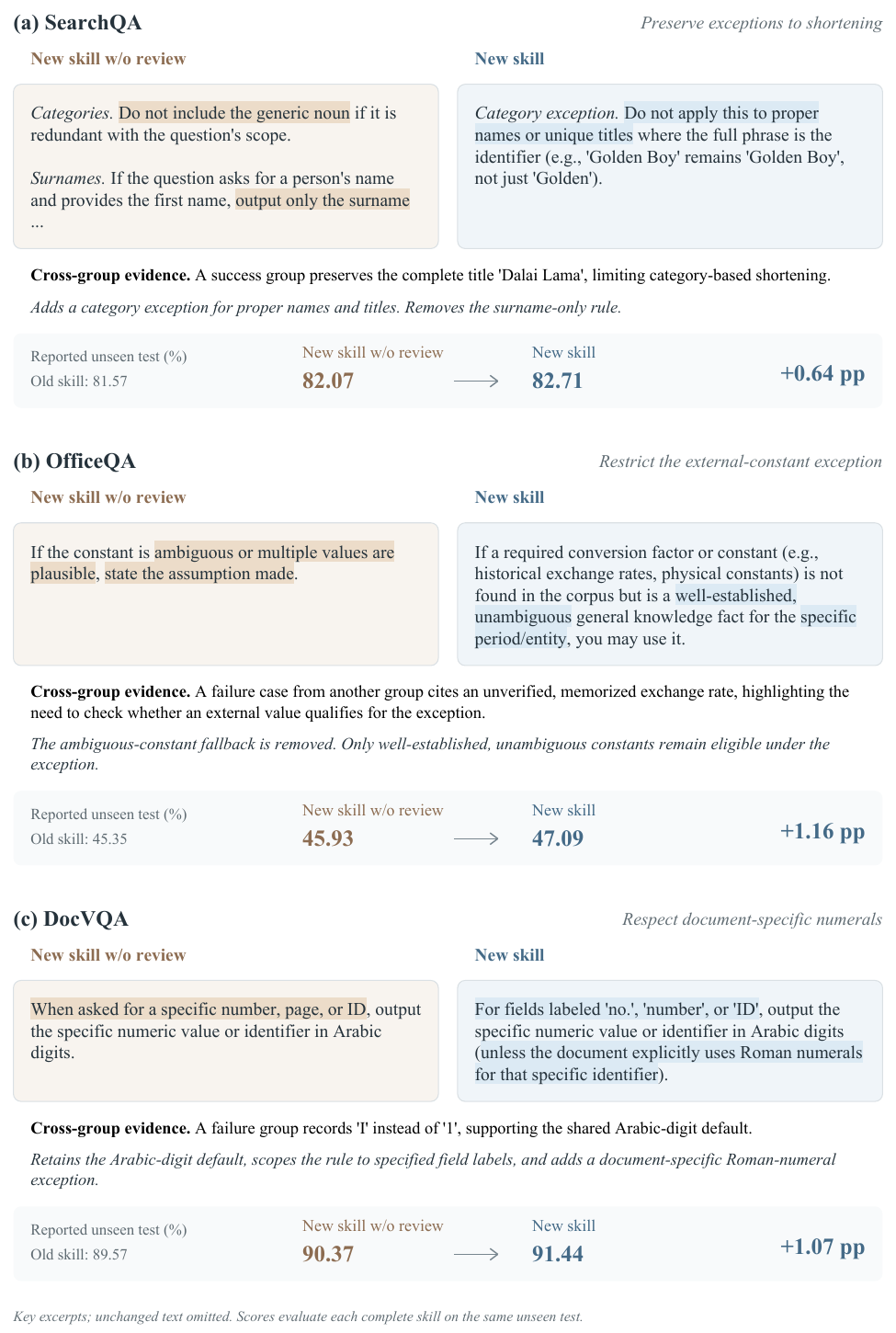}
    \caption{Key rule changes before and after evidence review for the updates in Table~\ref{tab:review-update-results}.}
    \label{fig:review-rule-changes}
\end{figure}

Figure~\ref{fig:review-rule-changes} illustrates how review refines the applicability of consolidated proposals. SearchQA protects proper names and titles during category extraction and removes the surname-only rule. OfficeQA removes the ambiguous-constant fallback while retaining a qualified exception for external constants. DocVQA preserves the Arabic-digit default, specifies the applicable field labels, and permits document-specific Roman numerals. In all three selected updates, review improves on an already beneficial proposal (Table~\ref{tab:review-update-results}) while clarifying the conditions and exceptions in the resulting guidance.

\clearpage

\clearpage
\section{Experimental results}
\label{app:experimental-results}

\subsection{Training Dynamics}
\label{app:training-dynamics}

\begin{figure}[!t]
    \centering
    \includegraphics[width=\linewidth]{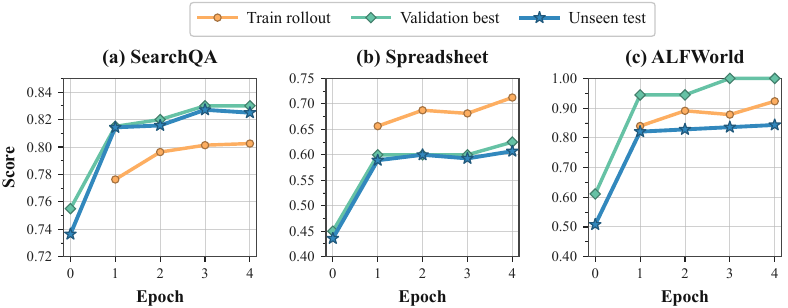}
    \caption{Epoch curves for TextualRL on SearchQA, Spreadsheet, and ALFWorld with Qwen3.8-27B. Epoch and score definitions follow Figure~\ref{fig:epoch-main}.}
    \label{fig:epoch-appendix}
\end{figure}

Epoch 0 denotes the Base Model. The training curve reports the mean rollout score within each epoch. Validation and test scores correspond to the best checkpoint selected by validation up to that epoch. Figure~\ref{fig:epoch-appendix} shows that SearchQA and Spreadsheet also obtain most of their test gains in the first epoch, increasing from 73.6\% to 81.4\% and from 43.6\% to 58.9\%, respectively. Subsequent changes are smaller, with final scores of 82.5\% and 60.7\%. ALFWorld's reported test score increases across all four epochs, from 50.8\% to 84.3\%, while validation reaches 100\% at epoch 3. Together with Figure~\ref{fig:epoch-main}, these curves show that the timing and magnitude of improvements vary by benchmark. A validation-selected checkpoint can remain unchanged across epochs, with repeated test evaluations yielding different scores.

\clearpage
\subsection{Optimized Contexts}
\label{app:skill-excerpts}

\begin{figure}[!t]
    \centering
    \includegraphics[width=\linewidth]{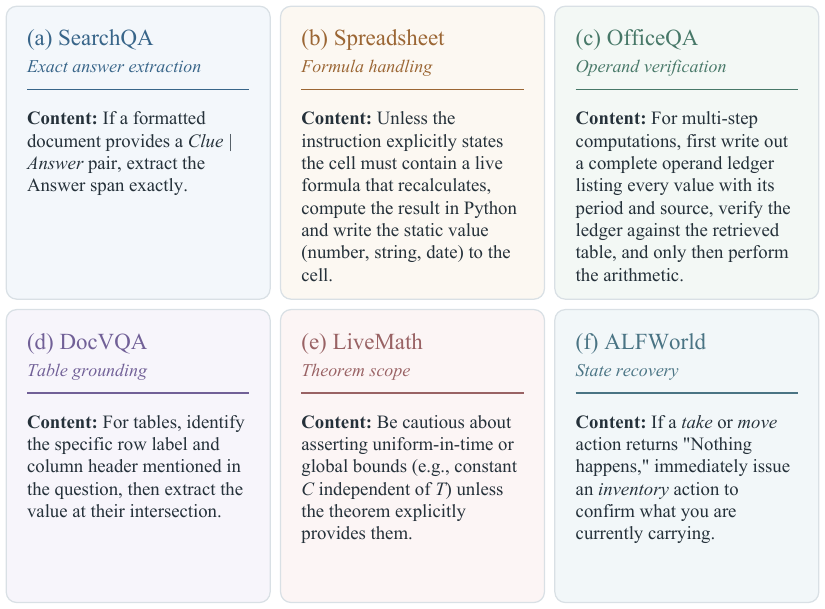}
    \caption{Representative instructions from the optimized contexts for six benchmarks. Original wording is preserved, with formatting standardized for readability.}
    \label{fig:skill-excerpts}
\end{figure}

To illustrate what TextualRL encodes in the optimized context, Figure~\ref{fig:skill-excerpts} presents one instruction excerpt from each of the six benchmarks for Qwen3.8-27B. These examples specify concrete operations, such as verifying source values before arithmetic and checking inventory after a failed action. They also define when a rule applies, including when to write static spreadsheet values and when a theorem supports a claimed bound.

\clearpage
\subsection{Optimizer Prompts}
\label{app:optimizer-prompts}

The boxes below reproduce the method-specific system-prompt blocks used by TextualRL, with line wrapping and formatting adjusted for readability. In the implementation, \emph{skill} denotes the editable context $C$. Each outcome-specific block defines the corresponding critique task, and all three routes share the evidence-card instruction. The final block guides proposal review against the pooled evidence.

\newtcolorbox{optimizerprompt}[1]{
    breakable,
    colback=textualrlblue!35!white,
    colframe=black!25,
    colbacktitle=textualrlblue,
    coltitle=black,
    boxrule=0.4pt,
    arc=1.5pt,
    left=7pt,right=7pt,top=6pt,bottom=6pt,
    fonttitle=\bfseries\small,
    fontupper=\small\raggedright,
    before skip=8pt,after skip=8pt,
    title={#1}
}

\subsubsection{Outcome Analysis}
The user message supplies the current context $C_t$, the edit budget, within-epoch history $H_t$, and the assigned task trajectories with their outcomes. The analyst returns a JSON object containing a proposed \texttt{patch} and \texttt{cross\_group\_cards}.

\begin{optimizerprompt}{Mixed outcomes: within-task contrast}
This call contains all unique trajectories from exactly one task whose K sibling rollouts include both successes and failures. Compare complete trajectories, locate the first observable behavioral divergence, and emit at most one atomic repair or preservation rule. The rule must be supported by the within-task contrast rather than by task literals or final evaluator labels. If no plausible causal divergence is visible, return an empty patch.
\end{optimizerprompt}

\begin{optimizerprompt}{All-success outcomes: cross-task preservation}
Each Task Block is one independent task whose sibling rollouts all succeeded. Multiple unique trajectories inside one Task Block are correlated alternatives, not independent votes. First compare Task Blocks semantically. Emit a rule only for an inference-time behavior shared by at least two distinct task\_ids. Treat the result as preservation evidence, not proof that the behavior caused success. If no cross-task invariant is visible, return an empty patch. Set \texttt{evidence\_kind} to \texttt{consensus} and \texttt{generality\_scope} to \texttt{cross\_task}.
\end{optimizerprompt}

\begin{optimizerprompt}{All-failure outcomes: cross-task repair}
Each Task Block is one independent task whose sibling rollouts all failed. Multiple unique trajectories inside one Task Block are correlated attempts, not independent votes. First cluster Task Blocks by a shared, observable failure mechanism. Emit a cautious repair hypothesis only when at least two distinct task\_ids expose the same mechanism. Do not infer a correct answer or task-specific procedure from failure alone. If no cross-task mechanism is visible, return an empty patch. Set \texttt{evidence\_kind} to \texttt{consensus} and \texttt{generality\_scope} to \texttt{cross\_task}.
\end{optimizerprompt}

\clearpage
\subsubsection{Evidence and Review}
The shared evidence instruction is included in all three critique routes. For cross-group review, the user message supplies $C_t$, the consolidated candidate patches, and the pooled current-step cards $E_t$. The review status \texttt{insufficient\_evidence} retains the proposal's original scope and uncertainty, as described in Section~\ref{sec:cross-group-evidence}.

\begin{optimizerprompt}{Shared instruction: trajectory evidence cards}
Keep the current route's proposal rules and edit budget. In the same JSON response, alongside patch, also return cross\_group\_cards, a list of objects: task\_id, rollout\_id, quote, condition, behavior, observed\_effect. For each displayed unique trajectory, report at most ONE salient behavioral observation, even when no edit is warranted (return patch with an empty edits list in that case). quote must be a short verbatim runtime action/observation span, up to 800 characters. Describe observable preconditions, the action, and its visible effect in at most 300 characters each. Use "unknown" when the trace does not establish a condition or effect. Do not invent a successful repair. Choose behavior relevant to a proposed repair or worth preserving. For mixed groups expose both sides of the divergence; for homogeneous groups describe the actual observed behavior, not a presumed cause of the final score. Only cite Task Block IDs and displayed trajectory IDs in this call. Do not cite gold answers, evaluator feedback, final grading labels, hidden siblings, or previous-step summaries as runtime evidence. Cards are evidence for later semantic comparison, not extra edits or independent votes. Do not put any card metadata or task IDs into executable skill content.
\end{optimizerprompt}

\begin{optimizerprompt}{Cross-Group Policy Update: evidence-based proposal review}
Alongside the candidate edits, you receive behavioral evidence from the current training step's mixed, all-success, and all-failure task groups. Associate each candidate ONLY with semantically relevant actions and observable preconditions. Benchmark names, task-ID similarity, and matching outcome labels are not evidence of relevance. Do not force all three group types to support every candidate.

Mixed evidence can suggest an improvement direction. Independent failure tasks can corroborate an observed failure mechanism, not prove a repair. Successful behavior can support preservation or LIMIT a repair's scope: do not prohibit a behavior merely because it also appears in a failed task. Distinguish a genuine contradiction under comparable conditions from different valid preconditions. Use an observable condition in executable content when narrowing a rule; keep supported exceptions and unrelated successful behavior intact. Only revise an existing proposed behavior; do not invent unrelated edits to use every card. If no relevant cross-group evidence exists, keep the original evidence scope and uncertainty rather than inventing support or discarding the edit solely for missing evidence. All-success and all-failure proposals remain eligible.

Maintain the normal output patch format and edit budget. For EACH output edit, add training-only cross\_group\_review with decision (retain, narrow, revise, or insufficient\_evidence), reason, and evidence (a list of task\_id, rollout\_id, relation [supports, limits, contradicts], reason). Cite only cards actually used. Correlated siblings are not independent task votes. These records do not automatically change support\_count, ranking, acceptance, or causal confidence. Keep metadata out of skill content. Existing protected sections and runtime observability restrictions still apply.
\end{optimizerprompt}

\Needspace{11\baselineskip}
\subsection{Complete Update Example}
\label{app:complete-update}

\begin{figure}[!t]
    \centering
    \includegraphics[width=\linewidth,height=0.94\textheight,keepaspectratio]{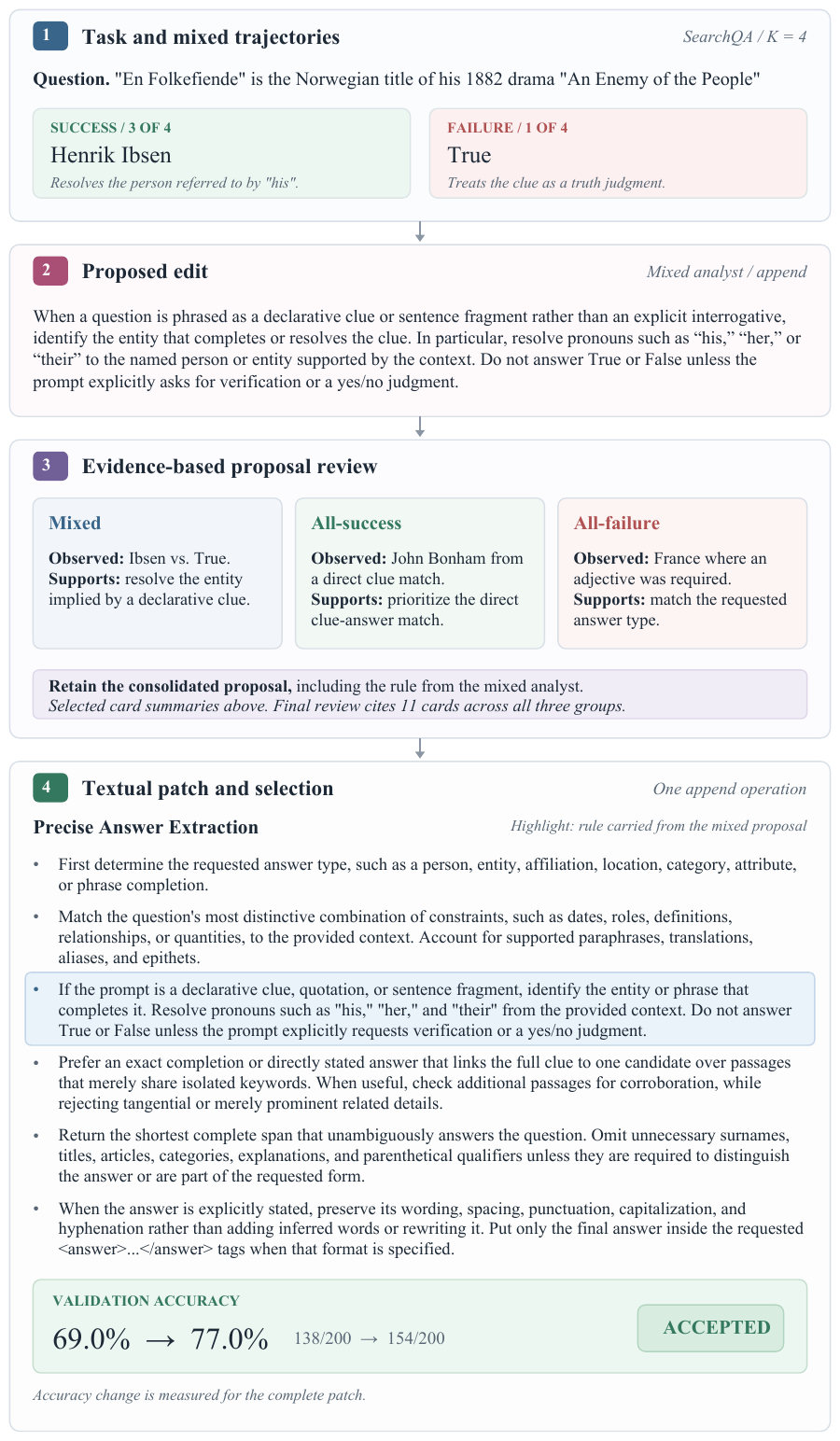}
    \caption{A complete context update example on SearchQA.}
    \label{fig:complete-update}
\end{figure}

Figure~\ref{fig:complete-update} traces an accepted SearchQA update using GPT-5.5 as both the frozen target and the optimizer. For the same declarative clue, three trajectories identify Henrik Ibsen, while one returns \texttt{True}. The mixed analyst proposes resolving the implied entity and reserving Boolean answers for explicit verification questions. The optimizer reviews the consolidated proposal against current-step evidence from all three outcome groups and retains this rule alongside complementary extraction guidance. The resulting patch is appended to a context with no learned rules and accepted after validation accuracy rises from 69.0\% to 77.0\%. This improvement is measured for the complete patch.

